\documentclass[twoside]{article}

\usepackage[accepted]{aistats2025}

\usepackage{amsmath,amssymb,amsfonts}
\usepackage{algorithmic}
\usepackage{graphicx}
\usepackage{textcomp}
\usepackage{xcolor}
\usepackage{microtype}
\usepackage{array}
\usepackage{multirow}
\usepackage{subcaption}
\usepackage{algorithm}
\usepackage{flushend}
\usepackage{booktabs} 
\usepackage{resizegather}
\usepackage{mathtools}
\usepackage{amsthm}
\usepackage{hyperref}
\usepackage{url}
\usepackage{caption}
\usepackage{bbm}
\usepackage{nccmath}

\usepackage[round]{natbib}

\begin{document}

%

%

\twocolumn[

\aistatstitle{Zero-Shot Action Generalization with Limited Observations}

\aistatsauthor{ Abdullah Alchihabi$^{1\ast}$ \And  Hanping Zhang$^{1\ast}$ \And Yuhong Guo$^{1,2}$ }

\aistatsaddress{ $^1$School of Computer Science, Carleton University, Ottawa, Canada  \\ 
$^2$Canada CIFAR AI Chair, Amii, Canada \\
\{abdullahalchihabi@cmail., jagzhang@cmail., yuhong.guo@\}carleton.ca}]

\begin{abstract}
Reinforcement Learning (RL) has demonstrated remarkable success in solving sequential decision-making problems. However, in real-world scenarios, RL agents often struggle to generalize when faced with unseen actions that were not encountered during training. Some previous works on zero-shot action generalization rely on large datasets of action observations to capture the behaviors of new actions, making them impractical for real-world applications. In this paper, we introduce a novel zero-shot framework, Action Generalization from Limited Observations (AGLO). Our framework has two main components: an action representation learning module and a policy learning module. The action representation learning module extracts discriminative embeddings of actions from limited observations, while the policy learning module leverages the learned action representations, along with augmented synthetic action representations, to learn a policy capable of handling tasks with unseen actions. The experimental results demonstrate that our framework significantly outperforms state-of-the-art methods for zero-shot action generalization across multiple benchmark tasks, showcasing its effectiveness in generalizing to new actions with minimal action observations.
\end{abstract}

\section{INTRODUCTION}

\begin{figure}[t]
\begin{subfigure}{0.235\textwidth}
\includegraphics[width = \textwidth]{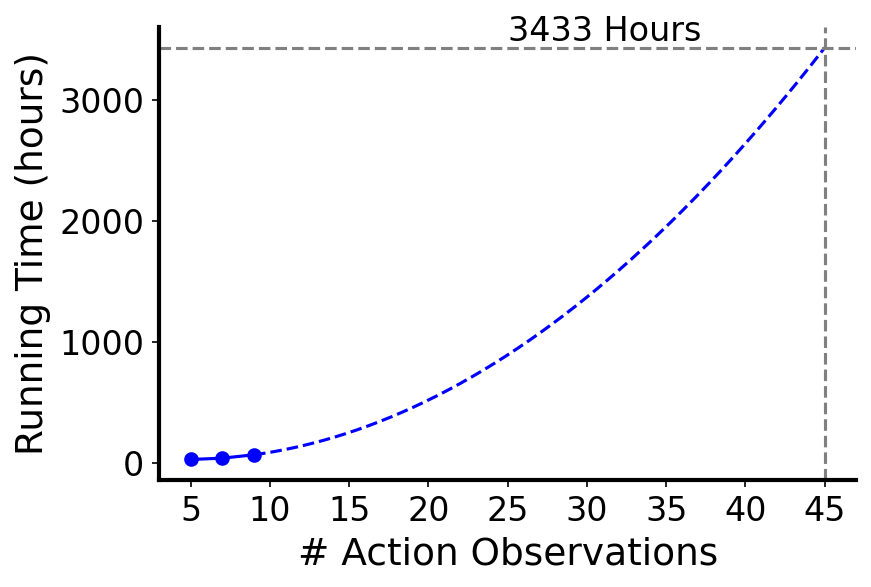} 
\caption{Observation Generation}
\label{fig:observation_generation}
\end{subfigure}
\begin{subfigure}{0.235\textwidth}
\centering
\includegraphics[width = \textwidth]{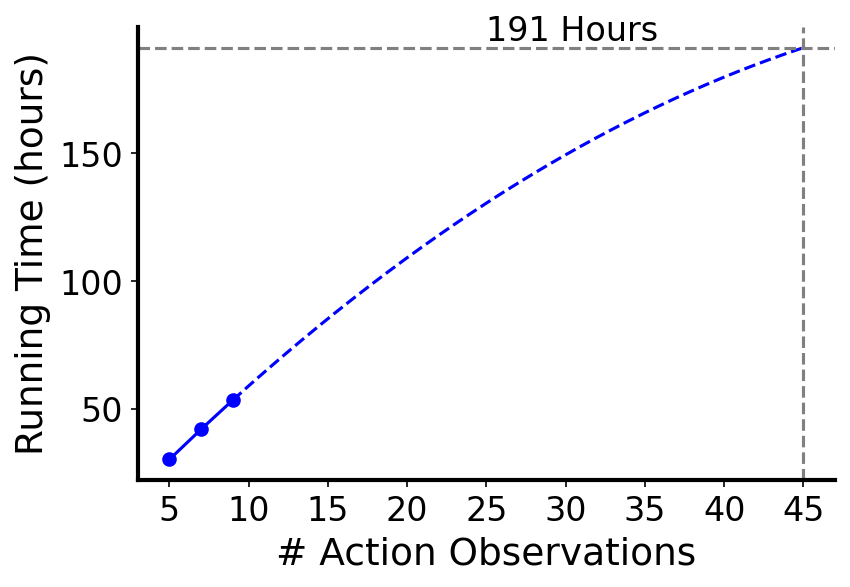}
\caption{Action Rep. Learning}
\label{fig:embedder_training}
\end{subfigure}
\caption{Estimated running time (in hours) for varying numbers of action observations in the CREATE environment. Previous studies utilized 45 observations per action \citep{jain2020generalization}. (a) Running time estimate for generating action observations. (b) Running time estimate for action representation learning.}
\label{fig:running_time}
\end{figure}

Reinforcement Learning (RL) has exhibited significant success in addressing sequential decision-making problems 
\citep{mnih2015human, silver2016mastering} across diverse application domains, from robotics \citep{kober2013reinforcement} to healthcare \citep{yu2021reinforcement}. 
However, this success is predominantly constrained to settings where agents are tested in environments identical or similar to those encountered during training, using the same set of actions. 
RL policies often exhibit poor generalization when faced with unseen tasks, novel environments, or unfamiliar action spaces. This limitation poses a significant challenge to deploying RL agents 
in dynamic, real-world settings. To mitigate this issue, recent work has sought to improve the generalization of RL agents, aiming to enhance their robustness and adaptability to new tasks and diverse environmental conditions \citep{cobbe2019quantifying, pinto2017robust, finn2017model}.
Despite these investigations, generalization to new
unseen actions at test time remains a relatively underexplored area. 
While several studies have investigated policy generalization to unseen actions \citep{chandak2020lifelong, ye2023action}, these approaches typically rely on fine-tuning the learned policies, which limits their applicability in real-world scenarios where retraining is impractical or costly. Recently, some work has focused on zero-shot generalization to unseen actions, where policies trained on observed actions are directly applied to new actions without any
fine-tuning or additional adaptation.
\citet{jain2020generalization} introduced a novel approach using an action embedder module, which learns action representations from task-agnostic action observations that capture the intrinsic properties of actions. These representations are then leveraged by a policy module for action selection. While this approach offers a promising step toward generalizing to unseen actions, it has a significant limitation. The action embedder requires a large number of observations for each action to generate informative and discriminative representations. In many real-world domains, obtaining such a large dataset of action observations is impractical, as these observations are typically generated through hand-crafted interactions between the action and environment by domain experts, a process that can be both time-consuming and expensive. 
Moreover, as the number of required observations increases, so does the time needed to generate the observations and to train the action embedder, as shown in Figure \ref{fig:running_time}. Figure \ref{fig:observation_generation} illustrates the relationship between the number of action observations and the 
time required for their generation in the CREATE environment, while Figure \ref{fig:embedder_training} highlights the time required to train the action embedder over the corresponding observations. This underscores the need for methods that can generalize to new, unseen actions without fine-tuning and with access to a very limited number of action observations. Developing such methods is crucial for enabling RL agents to be more broadly applicable in resource-constrained environments.

In response to these needs, in this work we propose a novel framework, named AGLO, for zero-shot Action Generalization from Limited Observations. 
This framework is made up of two modules: an action representation learning module and a policy learning module. The action representation learning module learns discriminative action embeddings (representations) from a limited number of associated observations. 
This is achieved by first using a coarse action observation encoder that learns a coarse representation for each action observation based solely on itself. 
Subsequently, a refined action observation encoder utilizes the similarity between coarse action observation representations and the corresponding action labels of each observation to learn refined action observations through graph contrastive learning and action observation representation classification. Furthermore, we employ a hierarchical variational auto-encoder to improve the learned observation representations through a 
reconstruction loss. 
Eventually, action representations are generated by pooling over the learned representations of their corresponding observations. 
The policy learning module leverages action representations to learn a dynamic policy. 
To improve the generalizability of the learned policy to previously unseen actions, we augment the action embedding space by generating synthetic action representations that encourage exploration during policy learning and minimize 
overfitting to actions encountered during training. 
We evaluate our proposed framework against state-of-the-art zero-shot action generalization methods on multiple tasks 
where each action is associated with a limited number of observations. The empirical results demonstrate the efficacy 
and superior generalization capacity of our proposed AGLO 
framework.

\section{RELATED WORKS}

\subsection{Generalization in Reinforcement Learning}

Generalization in RL is a fundamental challenge aimed at reducing the generalization gap between training and testing environments. The primary focus is on how well an RL policy learned during training can perform on previously unseen states. \citet{kirk2023survey} categorize existing methods to address generalization into three main types. 
The first type aims to increase the similarity between training and testing environments, 
leveraging techniques like data augmentation 
\citep{mazoure2022improving, zhang2021generalization} 
and environment generation \citep{jiang2021prioritized} to artificially enhance the diversity of training environments, thus improving the robustness of the learned policies across a broader range of testing conditions. 
The second type 
focuses on addressing discrepancies between the training and testing environments by modeling the variations and training policies that can adapt to such shifts \citep{song2019observational}. 
The third category focuses on RL-specific strategies, such as modifying value or policy functions to mitigate overfitting. Methods such as decoupling value and policy functions \citep{raileanu2021decoupling}, rethinking exploration strategies \citep{moon2022rethinking}, and fast adaptation techniques \citep{raileanu2020fast} have been proposed to improve policy generalization without requiring additional environment modifications. 
The aim of all these methods is to improve RL generalization and hence enable the deployment of RL in more robust, real-world applications where unseen states and environments are prevalent.

\subsection{Generalization to Unseen Actions}
Recently, the focus has shifted from generalization within the state space to the more challenging problem of handling unseen actions. While some prior works have investigated generalization to new 
actions, these approaches typically rely on fine-tuning the model for each new action \citep{chandak2020lifelong, ye2023action}, which limits their practicality in dynamic environments. 
\citet{jain2020generalization} introduced a pioneering method aimed at improving the performance of RL agents on unseen actions without the need for fine-tuning. Their approach leverages task-agnostic action observations to learn action embeddings through a Hierarchical Variational Autoencoder (HVAE) \citep{edwards2016towards}. This enables a policy learning module to select actions based on their learned embeddings, representing a significant step towards more adaptable 
RL agents. 
However, a key limitation of this method is its reliance on a large number of action observations to learn sufficiently representative action embeddings, which can be costly and time-consuming to obtain.
More recently, \citet{sinii2023context} proposed a Headless Algorithm Distillation method which employs a causal transformer to infer the embeddings of unseen actions based on in-context examples.
Despite these innovations, none of the existing approaches can generalize to unseen actions from a limited number of task-agnostic action observations without the need for policy fine-tuning, highlighting an open challenge in the field.

\section{PROBLEM SETUP}

In this work, we focus on the RL task of zero-shot generalization to new actions from a limited number of action observations. Specifically, the goal is to train a policy on a set of seen actions and evaluate its performance on a set of previously unseen actions without any re-training or fine-tuning on the new action set. Each action is associated with a small set of task-agnostic observations that capture the properties and behavior of the action, enabling the agent to generalize to unseen actions using only this limited information.

\subsection{Reinforcement Learning}

The RL problem is formulated as an episodic Markov Decision Process (MDP) with a discrete action space. 
Each episode of the MDP is characterized by a tuple $(\mathcal{S}, \mathcal{A}, \mathcal{T}, \mathcal{R}, \gamma)$, where $\mathcal{S}$ represents the set of states, $\mathcal{A}$ denotes the set of actions, $\mathcal{T}$ is the state transition function, $\mathcal{R}$ is the reward function, and $\gamma$ is the discount factor. 
During its interaction with the environment, at each timestep $t$, the RL agent selects an action $a_t \in \mathcal{A}$ using a policy $\pi$ 
based on the current state $s_t \in \mathcal{S}$.
The agent then receives a reward $\mathcal{R}(s_t, a_t)$ and transitions to a new state $s_{t+1}$ according to the environment’s transition dynamics $\mathcal{T}$. The objective of the agent is to maximize the 
accumulated discounted reward over the course of an episode, given by:
\begin{equation}
J = \sum_{t=0}^{T-1} \gamma^t r_t 
\end{equation}
where $T$ is the length of the episode
and $r_t = \mathcal{R}(s_t,a_t)$ denotes the reward at timestep $t$.

\subsection{Zero-Shot Action Generalization}

In the context of zero-shot generalization to new actions, the set of all actions is divided into two non-overlapping sets: the set of seen actions $\mathbb{A}$ and the set of unseen actions $\mathbb{A'}$. This setup consists of two sequential phases: the training phase and the evaluation phase. During the training phase, 
the set of actions available in each training episode $\mathcal{A}$ is sampled from the set of seen actions, $\mathbb{A}$. In the evaluation phase, the trained agent is tested in each evaluation episode using a set of actions $\mathcal{A}$ sampled from the set of unseen actions, $\mathbb{A'}$. 
The objective is to learn a policy $\pi(a|s,\mathcal{A})$ that maximizes the expected discounted reward for any action set $\mathcal{A}$ sampled from the set of unseen actions $\mathbb{A'}$: 
\begin{equation}
    J = \mathbb{E}_{\mathcal{A} \subset \mathbb{A'},\tau\sim\pi}  \Bigl[\sum\nolimits_{t=0}^{T-1} \gamma^{t} r_t\Bigr]
\end{equation}
where $\tau$ denotes a trajectory---a sequence of states, actions, and rewards---generated 
by interacting with environment using the policy.
Each action $a \in \mathbb{A} \cup \mathbb{A'}$ is associated with a set of 
observations $\mathcal{O} = \{o_1,\cdots,o_n\}$, where $n$ is the number of observations collected for each action $a$. These action observations can take the form of state trajectories, videos, or images that capture the intrinsic properties and behavior of the corresponding action. The set of observations associated with the seen actions is denoted by $\mathbb{O} = \{\mathcal{O}_1,\cdots,\mathcal{O}_N\}$, while the set of observations corresponding to the unseen actions is denoted by $\mathbb{O'} = \{\mathcal{O'}_1,\cdots,\mathcal{O'}_{N'}\}$, where $N$ and $N'$ represent the numbers of seen and unseen actions, respectively.

Previous works 
assume the availability of a large number of observations $\mathcal{O}$ for each action $a$, 
for both seen or unseen actions. 
However, this assumption introduces significant challenges in application domains where collecting a large number of action observations is expensive or impractical. Additionally, as the number of action observations increases, the cost and time required for generating such observations, as well as for training agents using these observations, increase considerably. Therefore, in this work, we operate under the assumption that the number of available observations for each action, $n$, is severely limited.

\section{PROPOSED METHOD}

In this section, we present our proposed AGLO framework for zero-shot Action Generalization with Limited Observations. 
The framework is composed of two key modules: the action representation learning module and the policy learning with action representation augmentation module. 
The action representation learning module is designed to generate discriminative action embeddings by encoding the limited set of observations for each action. 
The policy learning module leverages the learned action embeddings to augment the representations of the set of seen actions with synthetic action representations. 
The policy is subsequently trained using both the original representations of the actions and their augmented synthetic counterparts, facilitating the learning of a policy that generalizes effectively to unseen actions. 
In the following sections, we provide detailed descriptions of each module within our proposed AGLO framework.

\subsection{Action Representation Learning}

The action representation learning module is designed to generate representative action embeddings by encoding the limited set of observations associated with each action. To achieve this, the module first employs a coarse observation encoder, followed by a refined observation encoder that enhances cohesion and action discrimination in the learned observation representations. 
Subsequently, action representations are generated by pooling the associated refined observation embeddings. 
The action representations are further refined using a hierarchical variational auto-encoder to minimize the observation reconstruction loss. This ensures that the generated action embeddings faithfully capture the key characteristics of the action observations.

\subsubsection{Observation Representation Learning}

\paragraph{Coarse Action Observation Encoder} 

The coarse action observation encoder independently encodes each action observation, without considering the relationships between observations of the same action or other actions. Specifically, the coarse action observation encoder, denoted as $g_{\text{co}}$, takes an individual observation 
$o$ of an action $a$ as input, 
and outputs its coarse action observation embedding $c$ 
as follows: 
\begin{equation}
\label{eq:coarse}
    c = g_{\text{co}}(o).
\end{equation}

\paragraph{Refined Action Observation Encoder} 

The coarse action observation embeddings, generated independently, do not exploit the information available in other observations of the same action or in observations of other actions. Given the limited number of observations per action, it is crucial to leverage the 
available information across all observations to generate more representative action observation
embeddings. To achieve 
this, we propose a graph-based refined action observation encoder that encourages cohesion in the observation embedding space by promoting similarity between related observations while ensuring 
discrimination between different actions.

The input to the refined action observation encoder consists of a batch of actions and their corresponding observations. We begin by constructing an action observation graph $G = (V, E)$, where $V$ is the set of nodes, with each node $u$ representing an action observation $o$. The total number of nodes in the graph is $|V| = n \times K$, where $K$ denotes the number of actions in each batch. 
Each node $u$ is associated with an input feature vector that corresponds to the coarse observation embedding $c$ generated by the coarse action observation encoder $g_{\text{co}}$. The feature vectors of all nodes in the graph are organized into a matrix $C \in \mathbb{R}^{|V| \times d}$, where $d$ is the 
size of the coarse action observation embeddings. $E$ is the set of edges of the graph $G$, represented by a weighted undirected adjacency matrix $A$, which is constructed as follows:
\begin{equation}
\label{eq:Adjacency}
    A[u,v] =
\begin{cases}
    \sigma(C_u, C_v) & \text{if } \sigma(C_u, C_v) > \epsilon \\
    0 &  \text{otherwise}. 
\end{cases}
\end{equation}
Here, $\sigma(.,.)$ represents the Pearson correlation function, 
and  $\sigma(C_u, C_v)$
measures the similarity between the coarse action observation embeddings of nodes $u$ and $v$. The threshold $\epsilon$ controls the sparsity of the graph by filtering out edges with low similarities. 
Additionally, each node $u$ is associated with a label $y$ that indicates the action to which the corresponding observation belongs. The labels of all nodes are organized into a label indicator matrix $Y \in \{0, 1\}^{|V| \times K}$.

To effectively utilize the constructed action observation graph $G$, we define the refined observation encoder as a graph neural network-based function, $g_{\text{re}}$, that learns refined observation representations through message passing along the graph. The refined observation embeddings are generated as follows:
\begin{equation}
\label{eq:refined}
    \Tilde{C} = g_{\text{re}}(C,A).
\end{equation}
Here, $\Tilde{C}$ represents the refined action observation embedding matrix. To promote cohesion in the learned embedding space, we train both $g_{\text{re}}$ and $g_{\text{co}}$ to push 
similar observations closer to each other. 
This is achieved by minimizing the following contrastive loss:
\begin{equation}
\label{eq:cont-loss}
	\mathcal{L}_{\text{cont}} = -\!\sum_{u \in V}\! \log \frac{\exp(\sigma(\Tilde{C}_u, \Tilde{C}_{v_u}) / \kappa)}{\exp(\sigma(\Tilde{C}_u, \Tilde{C}_{v_u}) / \kappa) +\!\!\!\! 
	\sum\limits_{v' \in \mathcal{N}_u^c}\!\! \exp(\sigma(\Tilde{C}_u, \Tilde{C}_{v'}) / \kappa)}
\end{equation}
Here, node $v_u$ represents a positive instance randomly sampled from the set of nodes connected to node $u$ ($\mathcal{N}_u=\{ v \in V \mid A[u,v] > 0 \}$), while $\mathcal{N}_u^c$ denotes the set of negative instances of size $K'$ randomly sampled from the set of all nodes not connected to $u$ ($\mathcal{N}_u^c \subset V \setminus \mathcal{N}_u$). The temperature parameter $\kappa$ controls the scaling of the similarity scores. 
To further enhance action discrimination in the refined observation embeddings, we also define a classification function $\rho$ that predicts the corresponding action labels based on the refined embeddings 
as follows:
\begin{equation}
\label{eq:obs_class}
    P = \rho(\Tilde{C})
\end{equation}
where $P$ is the predicted action 
probability distribution. $g_{\text{re}}$, $g_{\text{co}}$, and $\rho$ are jointly trained to minimize the following action observation classification loss: 
\begin{equation}
\label{eq:ce_loss}
    \mathcal{L}_{ce} = \frac{1}{|V|} \sum\nolimits_{u \in V} \ell(P_u,Y_u)
\end{equation}
where $\ell$ denotes 
the standard cross-entropy loss function, 
and $P_u$ and $Y_u$ are the predicted and ground-truth 
class distribution vectors,
respectively, for node $u$.

\subsubsection{Action Representation Pooling}

The goal of the action representation learning module is to generate representative action embeddings by utilizing the refined observation embeddings for each action's observation set. Specifically, for each action $a_i$, the action representation function $g$ encodes the corresponding refined action observation embedding set 
$\Tilde{\mathcal{C}}_i = \{ \Tilde{C}_u  \mid u \in V \land Y_{u,i} = 1 \}$ as follows:
\begin{equation}
\label{eq:sample_action_gauss}
	(\mu_i , \sigma^2_i)   =  g(\Tilde{\mathcal{C}}_i)
\end{equation}
where the embedding of action $a_i$ is 
encoded
as a Gaussian distribution $\mathcal{N}(\mu_i, \sigma^2_i)$.

To ensure that the learned action embedding 
distributions effectively capture the information 
present in the corresponding action observations, we employ a Hierarchical Variational Auto-Encoder (HVAE) \citep{edwards2016towards}. The HVAE samples 
an action embedding 
$\hat{c}_{i}$ from the Gaussian distribution $\mathcal{N}(\mu_i, \sigma^2_i)$ and conditions both the observation encoder $q_{\psi}(z_{i}^{j}|o_{i}^{j}, \hat{c}_{i})$ and the decoder $p(o_{i}^{j}| z_{i}^{j}, \hat{c}_{i})$ on the
action embedding 
$\hat{c}_{i}$. This process
is applied over the entire set of observations $\mathcal{O}_i$ associated with action $a_i$, where $o_{i}^{j}$ is the j-th observation of action $a_i$ and $z_{i}^{j}$ represents the j-th encoded observation obtained from the action observation encoder $q_{\psi}$. 
The action representation function $g$ and the HVAE are trained jointly to minimize the following action observation reconstruction loss:
\begin{equation}
\label{eq:reconstruct_loss}
\begin{split}
	\mathcal{L}_{\text{reconst}} = &\sum_{\mathcal{O}_{i} \in \mathbb{O}} \biggl[ \mathbb{E}_{g(\hat{c}_{i}| \Tilde{\mathcal{C}}_i)}  \Bigl[ \sum_{o_{i}^{j} \in \mathcal{O}_{i}} 
    \mathbb{E}_{q_{\psi}(z_{i}^{j}|o_{i}^{j},\hat{c}_{i})} \log p(o_{i}^{j}|z_{i}^{j},\hat{c}_{i}) \\
    &- D_{\text{KL}}(q_{\psi}||p(z_{i}^{j}|\hat{c}_{i})) \Bigr] 
    - D_{\text{KL}}(g||p(\hat{c}_{i})) \biggr]
\end{split}
\end{equation}
where $D_{\text{KL}}$ represents the Kullback–Leibler (KL) divergence function.
The two $D_{\text{KL}}$  terms denote the KL-divergences 
between the output distributions of the observation encoder $q_\psi$ and its prior, 
and between the output distributions of the action representation function $g$ and its prior,
respectively.

All the components of the action representation learning module are trained jointly in an end-to-end fashion to minimize the following overall loss:
\begin{equation}
\label{eq:overall_loss}
   \mathcal{L}_{\text{total}} =  \mathcal{L}_{\text{reconst}} + \lambda_{\text{ce}} \mathcal{L}_{ce} + \lambda_{\text{cont}} \mathcal{L}_{\text{cont}}
\end{equation}
Here, $\lambda_{\text{ce}}$ and $\lambda_{\text{cont}}$ are hyper-parameters controlling the contributions of 
the observation classification loss $\mathcal{L}_{ce}$ and the contrastive learning loss $\mathcal{L}_{\text{cont}}$, respectively, to the overall loss.

\subsection{Policy Learning with Action Representation Augmentation}
\subsubsection{Action Representation Augmentation}

The policy learning module aims to learn a policy $\pi(a|s, \mathcal{A})$ over a sampled subset of seen actions $\mathcal{A} \subset \mathbb{A}$ that can generalize effectively to unseen actions $\mathbb{A'}$. The policy leverages the representations of these seen actions, which are produced by the action representation learning module, where each action representation $\hat{c}_i$ is 
sampled from a Gaussian distribution $\mathcal{N}(\mu_i, \sigma_i^2)$. 
To prevent the learned policy from overfitting to seen actions, we employ action representation augmentation to generate synthetic representations of the seen actions
in the learned action embedding space. These synthetic action representations promote exploration during policy learning, thereby reducing overfitting and enhancing the policy's generalization to unseen actions.

In particular, for each action $a_i$ with representation $\hat{c}_i$ sampled from a Gaussian distribution $\mathcal{N}(\mu_{i}, \sigma_{i}^{2})$, we randomly select another action $a_j \in \mathcal{A}$ ($a_i \neq a_j$) with representation $\hat{c}_j$ sampled from $\mathcal{N}(\mu_{j}, \sigma_{j}^{2})$. A new synthetic representation for $a_i$ is then generated by applying mixup augmentation in the action embedding space between $\hat{c}_i$ and $\hat{c}_j$, resulting in 
$\hat{c}_{i,\text{syn}}$
which follows a Gaussian distribution 
$\mathcal{N}(\mu_{i,\text{syn}}, \sigma_{i,\text{syn}}^{2})$, 
where the parameters $\mu_{i,\text{syn}}$ and $\sigma_{i,\text{syn}}^{2}$ are defined as follows:
\begin{equation}
\label{eq:mixup}
\begin{gathered}
\mu_{i,\text{syn}} = \lambda \mu_{i} + (1-\lambda) \mu_{j} \\
\sigma^{2}_{i,\text{syn}} = \lambda \sigma^{2}_{i}  + (1-\lambda) \sigma^{2}_{j} + \lambda (1-\lambda) ( \mu_{i}-\mu_{j})^{2}
\end{gathered}
\end{equation}
Here, $\lambda$ is the mixing coefficient sampled from a Beta distribution, 
$\text{Beta}(\alpha, \alpha)$, which controls the 
mixing rate between the two action embeddings. By applying this synthetic action representation generation process to each action $a \in \mathcal{A}$, we double the size of the action representation set in each episode where each action $a$ is represented by both its original representation generated based on its observations and a synthetic representation generated through augmentation. 

\subsubsection{Policy Learning}

The policy learning module consists of two key components: a state encoding function and a utility function.
The state encoding function, $f_{\omega}$, takes the current state $s_t$ as input and outputs a corresponding state encoding at time $t$. The utility function, $f_{\nu}$, then takes a representation $\hat{c}_i$ of an action $a_i$ and the state encoding $f_{\omega}(s_t)$ as input and outputs the predicted utility of the action $a_i$ in state $s_t$ based on the representation $\hat{c}_i$. 
Since each action $a_i$ is represented by both its original representation $\hat{c}_i$ 
and a synthetic representation $\hat{c}_{i,\text{syn}}$ generated through augmentation, 
the predicted score for $a_i$ is the sum of the utility scores predicted 
based on both $\hat{c}_i$ and $\hat{c}_{i,\text{syn}}$. 
The predicted scores for all actions in $\mathcal{A}$
are then normalized into a probability distribution using a softmax function as follows: 
\begin{equation}
\label{eq:action_selection}
\pi(a_i | s_t,\mathcal{A}) =\! 
	\frac{\exp({f_{\nu}(\hat{c}_{i},f_{\omega}(s_t))}) 
	\!+\!\exp({f_{\nu}(\hat{c}_{i,\text{syn}},f_{\omega}(s_t))}) 
	}
	{ \!\sum\limits_{j=1}^{K}\!\!\left(\exp(f_{\nu}(\hat{c}_{j},f_{\omega}(s_t)))
	\!+\!\exp(f_{\nu}(\hat{c}_{j,\text{syn}},f_{\omega}(s_t)))\right)
	}
\end{equation}
This allows the policy to select actions based on both original and synthetic action representations.

\begin{algorithm}[!t]
    \caption{Policy Training Procedure}
    \label{alg:policy_train}
    \begin{algorithmic}[1]
\STATE{\textbf{Input:} Seen actions $\mathbb{A}$, action observations $\mathbb{O}$, Learned Action Embedder \\ }
\STATE{\textbf{Output:} Learned Policy parameters }
\FOR{{iter = 1} {\bf to} maxiters}
\WHILE{episode not finished}
\STATE Sample a subset $\mathcal{A}$ of $K$ actions from $\mathbb{A}$
\STATE $\forall a \in \mathcal{A}, \hat{c} \sim \mathcal{N}(\mu,\sigma^2)$
	    \STATE Augment action representations via Eq.(\ref{eq:mixup})
	    \STATE Select action $a_t \sim \pi(\cdot|s_t,\mathcal{A})$ based on Eq.(\ref{eq:action_selection})
\STATE Execute action $a_t$: $s_{t+1}, r_t = \text{ENV}(s_t,a_t)$
\STATE Store $(s_t,a_t, r_t, s_{t+1})$ in replay buffer
\ENDWHILE
	    \STATE Update policy $\pi$ using Eq.(\ref{eq:policy_obj})
\ENDFOR
    \end{algorithmic}
\end{algorithm}

%
In some environments, the discrete actions are associated with additional parametrization intended to encode the location/coordinates of applying the selected action in the environment. In such cases, we define an auxiliary function $f_{\chi}$, which takes the state encoding as input and outputs the location/coordinates to apply the selected action, similar to \citep{jain2020generalization}. 
The overall action information in the environment is determined 
by sampling the action location/coordinates from the auxiliary function and the discrete action from Eq.(\ref{eq:action_selection}). 
The state encoding function, the utility function and the auxiliary function (if needed) are 
trained jointly in an end-to-end fashion using policy gradients \citep{sutton1999policy}. The objective of the policy training is to maximize the expected reward while 
promoting exploration. This is achieved by incorporating an entropy regularization term into the training objective, encouraging diversity in the actions selected by the policy:
\begin{equation}
\label{eq:policy_obj}
  \text{max}_{\nu, \omega, \chi}\mathbb{E}_{\mathcal{A} \subset \mathbb{A},\tau\sim\pi}  
	\bigg[\sum_{t=0}^{T-1}\gamma^tr_t +  \beta \mathcal{H}(\pi(\cdot|s_t,\mathcal{A})) \bigg]
\end{equation}
where $\mathcal{H}(.)$ denotes the entropy function, 
and $\beta$ is a hyper-parameter controlling the contribution of the entropy regularization term. 
The policy training procedure is presented in Algorithm \ref{alg:policy_train}.

\section{EXPERIMENTS}

\subsection{Experimental Setup}

\subsubsection{Environment}

We evaluate our AGLO framework in a sequential decision-making environment, Chain REAction Tool Environment (CREATE). 
CREATE \citep{jain2020generalization} is a physics-based environment where the agent's objective is to manipulate the trajectory of a ball in real-time by strategically placing tools in its path to guide the ball towards a goal position. 
In this work, we focus on three
tasks within this environment: Push, Navigate and Obstacle.
The agent's action selection involves choosing which tool to place and specifying the 2D coordinates for placing the selected tool. The action observations for each tool consist of state observations that describe the trajectories of a ball interacting with the tool at various speeds and from different directions. 
Further details about the CREATE environment can be found in \citep{jain2020generalization}. 

\begin{table*}[t]
\caption{The overall performance (standard deviation within brackets) on the Push, Navigate and Obstacle tasks in the CREATE environment with 3 different numbers of observations per action (5, 7 and 9).} 
\centering
\resizebox{\textwidth}{!}{ 
\begin{tabular}{l|l|l|l|l||l|l|l||l|l|l}
\hline
& & \multicolumn{3}{c||}{\textbf{Push Task}} &  \multicolumn{3}{c||}{\textbf{Navigate Task}} &  \multicolumn{3}{c}{\textbf{Obstacle Task}} \\
& {} & {Target Hit} & {Goal Hit} & {Reward} & {Target Hit} & {Goal Hit} & {Reward} & {Target Hit} & {Goal Hit} & {Reward} \\
\hline
\multirow{3}{*}{\rotatebox{90}{5 Obs.}} 
& {VAE} & ${25\%}_{(0.4)}$ & ${5\%}_{(0.1)}$ & ${0.80}_{(0.4)}$ 
 & ${1\%}_{(0.2)}$ & ${0\%}_{(0.0)}$ & ${0.10}_{(0.2)}$ 
 & ${2\%}_{(0.1)}$ & ${0\%}_{(0.0)}$ & ${0.83}_{(0.8)}$ \\
& {HVAE} & ${15\%}_{(0.8)}$ & ${3\%}_{(0.2)}$ & ${0.49}_{(0.2)}$
 & ${17\%}_{(0.1)}$ & ${2\%}_{(0.2)}$ & ${0.48}_{(0.3)}$ 
 & ${1\%}_{(0.1)}$ & ${0\%}_{(0.0)}$ & ${0.73}_{(0.8)}$ \\
& {AGLO} & $\mathbf{48\%}_{(0.1)}$ & $\mathbf{23\%}_{(0.1)}$ & $\mathbf{2.79}_{(0.1)}$
 & $\mathbf{49\%}_{(0.5)}$ & $\mathbf{13\%}_{(0.5)}$ & $\mathbf{1.82}_{(0.5)}$
 & $\mathbf{37\%}_{(0.1)}$ & $\mathbf{0\%}_{(0.0)}$ & $\mathbf{2.49}_{(0.2)}$ \\
\hline
\hline
\multirow{3}{*}{\rotatebox{90}{7 Obs.}} 
& {VAE} & ${15\%}_{(0.1)}$ & ${3\%}_{(0.3)}$ & ${0.55}_{(0.4)}$
& ${2\%}_{(0.2)}$ & ${1\%}_{(0.1)}$ & ${0.18}_{(0.1)}$
& ${1\%}_{(0.1)}$ & ${0\%}_{(0.0)}$ & ${0.82}_{(0.9)}$ \\
& {HVAE} & ${21\%}_{(0.3)}$ & ${3\%}_{(0.1)}$ & ${0.58}_{(0.5)}$
 & ${2\%}_{(0.2)}$ & ${1\%}_{(0.1)}$ & ${0.13}_{(0.7)}$
 & ${1\%}_{(0.1)}$ & ${1\%}_{(0.1)}$ & ${1.71}_{(0.1)}$ \\
& {AGLO} & $\mathbf{46\%}_{(0.3)}$ & $\mathbf{20\%}_{(0.2)}$ & $\mathbf{2.42}_{(0.2)}$
& $\mathbf{44\%}_{(0.5)}$ & $\mathbf{11\%}_{(0.3)}$ & $\mathbf{1.56}_{(0.3)}$
& $\mathbf{34\%}_{(0.4)}$ & $\mathbf{7\%}_{(0.1)}$ & $\mathbf{3.03}_{(0.8)}$ \\
\hline
\hline
\multirow{3}{*}{\rotatebox{90}{9 Obs.}} 
& {VAE} & ${30\%}_{(0.1)}$ & ${6\%}_{(0.3)}$ & ${0.96}_{(0.3)}$
& ${8\%}_{(0.1)}$ & ${1\%}_{(0.1)}$ & ${0.26}_{(0.2)}$
& ${8\%}_{(0.1)}$ & ${0\%}_{(0.0)}$ & ${1.13}_{(1.2)}$ \\
& {HVAE} & ${25\%}_{(0.3)}$ & ${5\%}_{(0.1)}$ & ${0.79}_{(0.1)}$
 & ${1\%}_{(0.1)}$ & ${0\%}_{(0.0)}$ & ${0.09}_{(0.1)}$
 & ${1\%}_{(0.1)}$ & ${0\%}_{(0.0)}$ & ${0.16}_{(0.1)}$ \\
& {AGLO} & $\mathbf{47\%}_{(0.1)}$ & $\mathbf{21\%}_{(0.4)}$ & $\mathbf{2.52}_{(0.4)}$
& $\mathbf{50\%}_{(0.1)}$ & $\mathbf{16\%}_{(0.1)}$ & $\mathbf{2.05}_{(0.1)}$
& $\mathbf{33\%}_{(0.4)}$ & $\mathbf{8\%}_{(0.1)}$ & $\mathbf{3.19}_{(1.1)}$ \\
\hline
\end{tabular}}
\label{table:combined_task_results}
\end{table*}

\subsubsection{Evaluation Setup}

We follow the evaluation procedure outlined in \citep{jain2020generalization}, where actions are divided into three non-overlapping sets: train, validation, and test actions. 
We use the same train/validation/test action split
provided by \citet{jain2020generalization}, where 50\% of the actions are allocated to the training set, 
25\% to the validation set, and 25\% to the test set.
For each action, a limited number of observations is generated. The action representation learning module is trained on the training actions set and their corresponding action observations. Once trained, the action representation learning module is used to generate embeddings for the actions in the training set, and a policy is trained to solve the environment tasks using these learned action representations.

During the evaluation phase, the learned policy is assessed on the test action set over 200 episodes, where the action set for each episode is randomly sampled from the test set. The representations of test actions are inferred from their corresponding observations using the action representation learning module. We report the mean and standard deviation across three runs for the policy learning module, each with 200 evaluation episodes, for three key evaluation metrics: episode target hit, episode goal hit, and episode reward. 
The target hit metric indicates whether the target ball moved, while the goal hit metric reflects whether the goal was reached.

\subsection{Implementation Details}

We utilize the $g_{\text{co}}$ and $g$ implementations provided in \citep{jain2020generalization}, along with the HVAE implementation from \citep{edwards2016towards}, with an action embedding size of 128. The message passing function $g_{\text{re}}$ consists of two Graph Convolution Network layers \citep{kipf2016semi}, each followed by a ReLU activation function. The sparsity threshold ($\epsilon$) of the action observation adjacency matrix $A$ is set to $0.95$. 
The action observation classifier $\rho$ 
is composed of two fully connected layers, each  followed by a ReLU activation function. The contrastive loss $\mathcal{L}_{\text{cont}}$ uses a temperature value of $\kappa = 0.5$, a negative observation set size of $K' = 5$, and a hyper-parameter $\lambda_{\text{cont}}$ with a value of $1e^{-1}$. 
The hyper-parameter $\lambda_{ce}$ for the observation classification loss is set to $1e^{-3}$. 
The action representation learning module is trained using the RAdam optimizer with a learning rate of $1e^{-3}$, 
over 10,000 epochs, and a batch size of $K = 32$.

For the policy learning module, 
we adopt the implementations of the state encoder, utility function, and auxiliary function 
from \citep{jain2020generalization}. 
We utilize the Proximal Policy Optimization (PPO) method \citep{schulman2017proximal}  
to learn the policy function $\pi$, 
and extend the specific PPO implementation from \citep{jain2020generalization}. 
The number of sampled actions is set to $K=50$ and the maximum episode length is set to 30 timesteps. 
The mixing rate $\lambda$ for generating augmented actions representations is sampled 
from a Beta($0.4, 0.4$) distribution.
The policy is trained using the Adam optimizer 
with a learning rate of $1e^{-3}$ for $6e^{7}$ steps 
and a batch size of 3,072. 
The entropy coefficient $\beta$ is set to $5e^{-3}$. 

\subsection{Comparison Results}

We evaluate the performance of our AGLO framework on the zero-shot action generalization problem with limited action observations. Specifically, we conduct experiments using three different numbers of observations per action: 5, 7 and 9 observations. These correspond to 10\%, 15\% and 20\% of the observations per action used in prior work \citep{jain2020generalization}. 
We compare our AGLO against two zero-shot action generalization methods: Hierarchical Variational Auto-Encoder (HVAE) \citep{jain2020generalization} and Variational Auto-Encoder (VAE) \citep{jain2020generalization}. The comparison results for the Push, Navigate and Obstacle tasks are presented in Table \ref{table:combined_task_results}.

Table \ref{table:combined_task_results} shows 
that our AGLO framework consistently and significantly outperforms both the VAE and HVAE methods across all three tasks, for all three numbers of action observations, and across all evaluation metrics.
The performance improvements on the Push task are substantial, with gains exceeding 23\%, 25\% and 17\% in terms of Target Hit, and 18\%, 17\% and 15\% in terms of Goal Hit for 5, 7 and 9 action observations, respectively. 
Moreover, AGLO demonstrates superior performance in terms of average rewards collected on the Push task, 
achieving remarkable improvements of approximately 1.99 (248\% relative gain), 1.84 (317\% relative gain) and 1.56 (162\% relative gain) with 5, 7 and 9 action observations, respectively. 
The performance gains on the more challenging Navigate and Obstacle tasks are even more pronounced, where both VAE and HVAE struggle to learn generalizable policies. 
On the Navigate task, AGLO achieves significant gains in terms of Target Hit with improvements of around 32\%, 42\% and 42\% for 5, 7 and 9 observations, respectively. 
Similarly, on the Obstacle task, AGLO outperforms both VAE and HAVE in 
terms of Target Hit by 35\%, 33\% and 25\% for 5, 7 and 9 action observations, respectively. 
These results highlight the superior performance of our framework compared to existing zero-shot action generalization methods, particularly in scenarios where the number of available action observations is limited, across three challenging sequential decision-making tasks.

\subsection{Ablation Study}

We conducted an ablation study to investigate the contributions of 
three novel components in our framework:
the observation action classification loss $\mathcal{L}_{ce}$ 
and the action observation contrastive learning loss $\mathcal{L}_{\text{cont}}$
in the action representation learning module, 
as well as
the action representation augmentation (denoted as Action Aug.) in the policy learning module.
Specifically, we consider five variants of our proposed AGLO framework, 
where in each variant we remove one or two of the three novel components. 
The evaluation is carried out on the CREATE Push task with 5 observations per action, and the results are presented in Table \ref{table:ablation}.

The results in Table \ref{table:ablation} demonstrate 
performance degradations in all variants across the three evaluation metrics compared to the full proposed framework. 
The significant performance drops are observed in variants that do not include action representation augmentation,
highlighting the critical role of action representation augmentation 
in improving the generalization of the learned policy.  
Moreover, the notable performance drops when $\mathcal{L}_{\text{cont}}$ and $\mathcal{L}_{ce}$ are removed underscore the importance of both of these losses in learning discriminative and cohesive action representations, particularly as reflected in the reward and goal hit metrics. 
The consistent performance degradations across all metrics for all variants 
highlight the essential contribution of each novel component in our proposed framework.

\subsection{Analysis of Policy Learning}

\begin{figure*}[!t]
\centering
\includegraphics[width=0.75\textwidth]{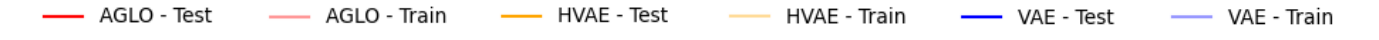}\\
\begin{subfigure}{0.24\textwidth}
\centering
\includegraphics[width = \textwidth]{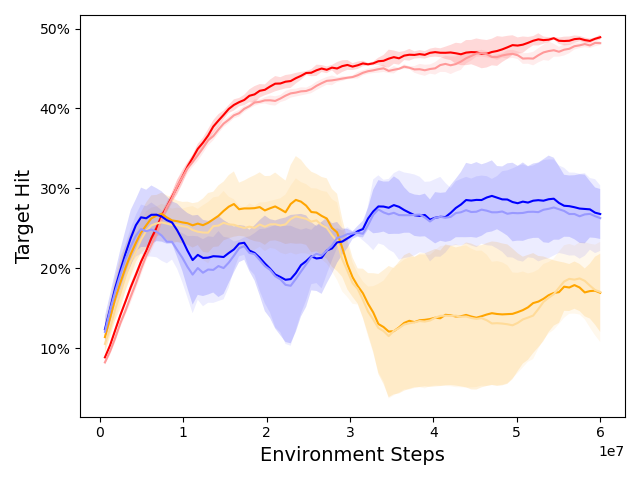} 
\caption{5 Observations, Target Hit}
\label{fig:push_target_10}
\end{subfigure}
\begin{subfigure}{0.24\textwidth}
\centering
\includegraphics[width = \textwidth]{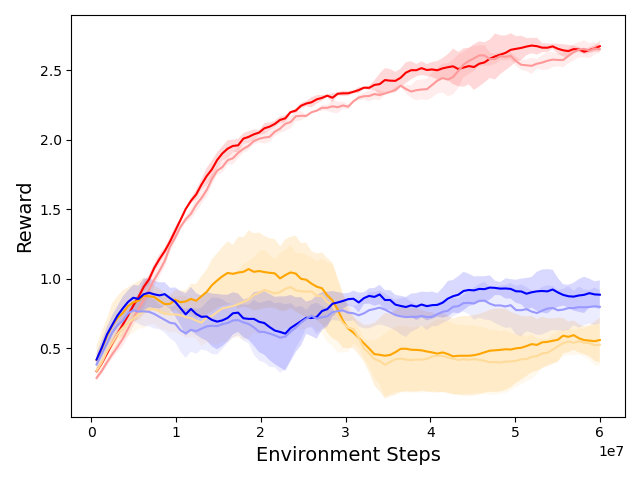}
\caption{5 Observations, Reward}
\label{fig:push_reward_10}
\end{subfigure}
\begin{subfigure}{0.24\textwidth}
\centering
\includegraphics[width = \textwidth]{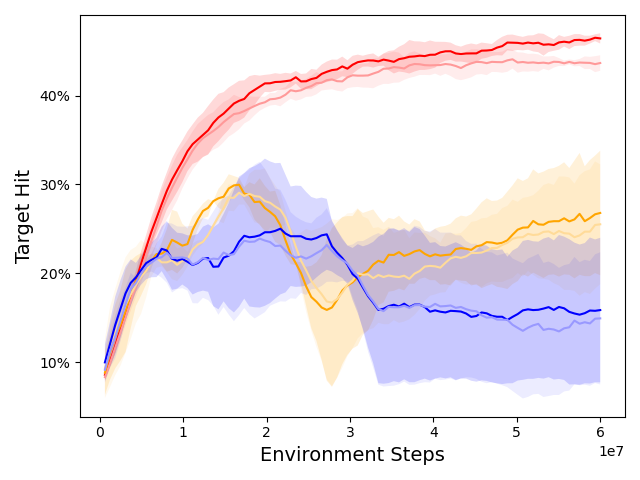}
\caption{7 Observations, Target Hit}
\label{fig:push_target_15}
\end{subfigure}
\begin{subfigure}{0.24\textwidth}
\centering
\includegraphics[width = \textwidth]{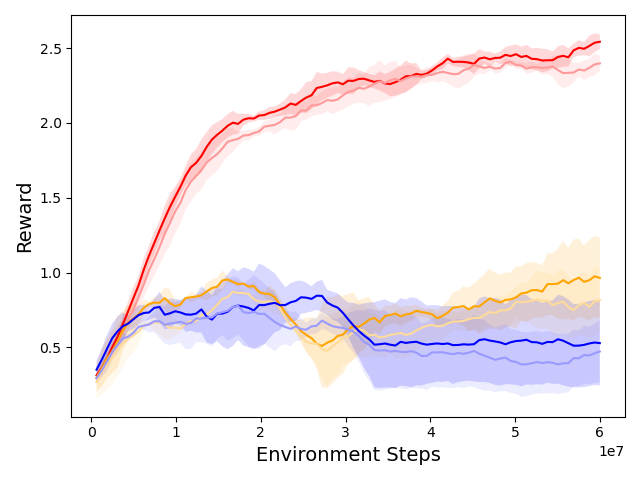} 
\caption{7 Observations, Reward}
\label{fig:push_reward_15}

\end{subfigure}
\caption{Policy learning analysis of VAE, HVAE and our AGLO on the Push task, with 5 and 7 observations per action, 
	evaluated using the Target hit and reward metrics.}
\label{fig:curve}
\end{figure*}

We also investigate the dynamics of the policy learning process within our AGLO framework. Specifically, we sample 200 training and 200 testing episodes, where the actions in the training episodes are drawn from the set of training actions, and the actions in the testing episodes are drawn from the set of testing actions. 
We evaluate the performance of our framework on both sets of episodes throughout the policy learning process by measuring the mean and standard deviation of Target hit and collected rewards across three runs. To assess its robustness, we compare the performance of our AGLO framework against both VAE and HVAE methods on the Push task with 5 and 7 observations per action, and present the results in Figure \ref{fig:curve}.

Figure \ref{fig:curve} demonstrates that while all three methods perform similarly in the early stage of policy learning, the performance of AGLO framework diverges significantly from that of VAE and HVAE as learning progresses. 
In particular, AGLO exhibits steady and consistent improvement in terms of both rewards and Target hit metrics as the number of environment steps increases, across both numbers of action observations. This improvement is observed in both the training and testing episodes, highlighting that our framework effectively learns stable and generalizable policies that perform well on both seen and unseen actions. 
In contrast, HVAE and VAE methods either converge to suboptimal policy solutions—indicative of a lack of exploration during learning—or develop unstable policies where performance fluctuates or drops significantly during 
policy learning. 
Our AGLO framework also exhibits smaller standard deviations across different runs compared to both 
the HVAE and VAE methods, highlighting the robustness of its training process. 
These results underscore the superior performance of AGLO in learning stable policies that generalize well to unseen actions, while also highlighting the limitations of current methods.

\begin{table}[t!]
\caption{Ablation study results (standard deviation within brackets) on the Push task in the CREATE environment with 5 observations per action.}
\centering
\setlength{\tabcolsep}{5.0pt} 
\resizebox{\columnwidth}{!}{ 
\begin{tabular}
{p{0.5cm}|p{0.62cm}|p{0.9cm}|c|c|c}
\hline
 \centering $\mathcal{L}_{{ce}}$& \centering $\mathcal{L}_{\text{cont}}$& \centering  Action\newline Aug. & {Target Hit}  & {Goal Hit} & {Reward} 
 \\ 
\hline 
\centering \checkmark & \centering \checkmark & \centering \checkmark & $\mathbf{48\%}_{(0.1)}$   & $\mathbf{23\%}_{(0.1)}$  & $\mathbf{2.79}_{(0.1)}$    \\
\centering \checkmark & \centering  \checkmark &    & ${27\%}_{(0.8)}$   & ${7\%}_{(0.1)}$   & ${1.03}_{(0.1)}$  \\
& \centering \checkmark & \centering  \checkmark &  ${46\%}_{(0.3)}$ & ${19\%}_{(0.2)}$ & ${2.37}_{(0.2)}$    \\
\centering  \checkmark &  &  \centering \checkmark &  $\mathbf{48\%}_{(0.1)}$ & ${21\%}_{(0.1)}$ & ${2.53}_{(0.1)}$  \\
& \centering \checkmark &  & ${21\%}_{(0.1)}$   & ${6\%}_{(0.1)}$  & ${0.82}_{(0.5)}$   \\
\centering  \checkmark &  &   & ${26\%}_{(0.3)}$ & ${7\%}_{(0.2)}$ & ${0.98}_{(0.2)}$  \\
\hline
\end{tabular}
	}
\label{table:ablation}
\end{table}

\section{CONCLUSION}

In this paper, we introduced a novel AGLO framework to tackle the challenging problem 
of zero-shot action generalization in Reinforcement Learning, specifically targeting scenarios with limited action observations. Our proposed framework consists of two key components: an action representation learning module and a policy learning module.
The action representation learning module encodes the limited set of action observations to generate representative action embeddings. 
The policy learning module then leverages the learned action embeddings to augment the representations of seen actions with synthetic action representations. 
The policy is subsequently trained on both the original representations of seen actions and their 
synthetic counterparts, enabling the learning of a policy that generalizes effectively to unseen actions. 
Our extensive experiments 
on three benchmark tasks demonstrate the superior zero-shot generalization capabilities of the policies learned by our proposed AGLO framework 
when applied to unseen actions with limited observations.

\bibliography{egbib}

\end{document}